\documentclass[runningheads]{llncs}
\usepackage{graphicx}
\usepackage{url}            
\usepackage{booktabs}       
\usepackage{amsmath}
\usepackage{amssymb}
\usepackage{amsfonts}       
\usepackage{nicefrac}       
\usepackage{microtype}      
\usepackage{tikz}
\usepackage{tabularx}
\usepackage{hhline}
\usepackage{adjustbox}
\usepackage{multirow}
\usepackage{mathtools}
\usepackage{bbm}
\usepackage{bm}
\usepackage{color, colortbl}
\usepackage{pifont}
\usepackage{appendix}
\usepackage{diagbox}
\usepackage[misc]{ifsym}

\usepackage[colorlinks,linkcolor=red, anchorcolor=blue, citecolor=teal, urlcolor=magenta]{hyperref}

\usepackage{cellspace}
\DeclarePairedDelimiterX\set[1]\{\}{\nonscript\,#1\nonscript\,}

\usetikzlibrary{shapes.geometric, arrows}
\tikzstyle{startstop} = [rectangle, rounded corners, minimum width=1.5cm, minimum height=0.8cm,text centered, draw=black, fill=red!30]
\tikzstyle{io} = [trapezium, trapezium left angle=70, trapezium right angle=110, minimum width=1.5cm, minimum height=0.8cm, text centered, draw=black, fill=blue!30]
\tikzstyle{process} = [rectangle, minimum width=1.5cm, minimum height=0.8cm, text centered, draw=black, fill=orange!30]
\tikzstyle{decision} = [diamond, minimum width=1.5cm, minimum height=0.8cm, text centered, draw=black, fill=green!30]
\tikzstyle{block} = [rectangle, rounded corners, minimum width=1.5cm, minimum height=0.8cm,text centered, draw=black, fill=orange!30]
\tikzstyle{network} = [rectangle, rounded corners, minimum width=1.5cm, minimum height=0.8cm,text centered, draw=black, fill=green!30]
\tikzstyle{bim} = [rectangle, rounded corners, minimum width=0.1cm, minimum height=0.1cm, text centered, draw=blue, dash pattern=on 4pt off 4pt]
\tikzstyle{arrow} = [thick,->,>=stealth]

\usepackage{makecell}

\newcommand\tf[1]{\textbf{#1}}

\def\ie{\textit{i.e.}}

\newcommand{\myparagraph}[1]{\vspace{1pt}\noindent{\bf{#1}}~~}
\makeatletter
\renewcommand{\paragraph}{%
  \@startsection{paragraph}{4}%
  {\z@}{0em}{-1em}%
  {\normalfont\normalsize\bfseries}%
}
\makeatother

\usepackage{colortbl}
\definecolor{lightgray}{gray}{0.75}
\definecolor{lightergray}{gray}{0.85}
\definecolor{Blue}{RGB}{3, 31, 97}
\definecolor{Blue1}{RGB}{214, 235, 245}
\definecolor{Blue2}{RGB}{235, 245, 250}
\definecolor{Gray}{RGB}{247, 252, 255}

\definecolor{convcolor}{HTML}{412F8A}
\definecolor{resnetcolor}{HTML}{8DA0CB}
\definecolor{vitcolor}{HTML}{fc8e62}

\definecolor{aliceblue}{rgb}{0.94, 0.97, 1.0}

\newcommand{\bs}[1]{\boldsymbol{#1}}

\begin{document}
\title{Implicit Anatomical Rendering for Medical Image Segmentation with Stochastic Experts}
\titlerunning{Implicit Anatomical Rendering for Medical Image Segmentation}
%
\author{Chenyu You \textsuperscript{1(\Letter)} \and Weicheng Dai \inst{2} \and Yifei Min \inst{4} \and Lawrence Staib \inst{1,2,3} \and \\ James S. Duncan \inst{1,2,3,4}}
\authorrunning{C. You et al.}
\institute{\textsuperscript{1}Department of Electrical Engineering, Yale University
\\
\email{chenyu.you@yale.edu}\\
\textsuperscript{2}Department of Radiology and Biomedical Imaging, Yale University\\
\textsuperscript{3}Department of Biomedical Engineering, Yale University \\
\textsuperscript{4}Department of Statistics and Data Science, Yale University\\
}

\maketitle              
\begin{abstract}
Integrating high-level semantically correlated contents and low-level anatomical features is of central importance in medical image segmentation. Towards this end, recent deep learning-based medical segmentation methods have shown great promise in better modeling such information. However, convolution operators for medical segmentation typically operate on regular grids, which inherently blur the high-frequency regions, \ie, boundary regions. In this work, we propose MORSE, a generic implicit neural rendering framework designed at an anatomical level to assist learning in medical image segmentation. Our method is motivated by the fact that implicit neural representation has been shown to be more effective in fitting complex signals and solving computer graphics problems than discrete grid-based representation. The core of our approach is to formulate medical image segmentation as a rendering problem in an end-to-end manner. Specifically, we continuously align the coarse segmentation prediction with the ambiguous coordinate-based point representations and aggregate these features to adaptively refine the boundary region. To parallelly optimize multi-scale pixel-level features, we leverage the idea from Mixture-of-Expert (MoE) to design and train our MORSE with a stochastic gating mechanism. Our experiments demonstrate that MORSE can work well with different medical segmentation backbones, consistently achieving competitive performance improvements in both 2D and 3D supervised medical segmentation methods. We also theoretically analyze the superiority of MORSE.

\end{abstract}
\section{Introduction}
\label{section:intro}

Medical image segmentation is one of the most fundamental and challenging tasks in medical image analysis. 
It aims at classifying each pixel in the image into an anatomical category. With the success of deep neural networks (DNNs), medical image segmentation has achieved great progress in assisting radiologists in contributing to a better disease diagnosis.  

Until recently, the field of medical image segmentation has mainly been dominated by an encoder-decoder architecture, and the existing state-of-the-art (SOTA) medical segmentation models are roughly categorized into two groups: (1) convolutional neural networks (CNNs) \cite{ronneberger2015u,cciccek20163d,oktay2018attention,chen2020realistic,xue2018segan,zhou2018unetplus,lai2022sar,lai2022smoothed,he2021interpretable,you2020unsupervised,you2022bootstrapping}, and (2) Transformers\cite{chen2021transunet,hatamizadeh2021unetr,you2022class}. 
However, despite their recent success, several challenges persist to build a robust medical segmentation model: \ding{182} Classical deep learning methods require precise pixel/voxel-level labels to tackle this problem \cite{you2022simcvd,you2021momentum,you2022mine,you2023rethinking,you2023action++}. Acquiring a large-scale medical dataset with exact pixel- and voxel-level annotations is usually expensive and time-consuming as it requires extensive clinical expertise \cite{lin2022cascade,lai2021joint,lai2021semi,oliveira2022generalizable,lai2022brainsec,you2020unsupervised}. 
Prior works \cite{li2020contrastive,huang2022boundary} have used point-level supervision on medical image segmentation to refine the boundary prediction, where such supervision requires well-trained model weights and can only capture discrete representations on the pixel-level grids. 
\ding{183}  Empirically, it has been observed that CNNs inherently store the discrete signal values in a grid of pixels or voxels, which naturally blur the high-frequency anatomical regions, \ie, boundary regions. 
In contrast, implicit neural representations (INRs), also known as coordinate-based neural representations, are capable of representing discrete data as instances of a continuous manifold, and have shown remarkable promise in computer vision and graphics \cite{park2019deepsdf,sitzmann2020implicit,tancik2020fourier}. 
Several questions then arise: \textit{how many pixel- or voxel-level labels are needed to achieve good performance? how should those coordinate locations be selected? and how can the selected coordinates and signal values be leveraged efficiently?}

Orthogonally to the popular belief that the model architecture matters the most in medical segmentation (\ie, complex architectures generally perform better), this paper focuses on an under-explored and alternative direction: \textit{towards improving segmentation quality via rectifying uncertain coarse predictions.} 
To this end, we propose a new INR-based framework, MORSE (i\underline{\tf{M}}plicit anat\underline{\tf{O}}mical \underline{\tf{R}}endering with \underline{\tf{S}}tochastic \underline{\tf{E}}xperts). 
The core of our approach is to formulate medical image segmentation as a rendering problem in an end-to-end manner.
We think of building a generic implicit neural rendering framework to have fine-grained control of segmentation quality, \ie, to adaptively compose coordinate-wise point features and rectify uncertain anatomical regions. 
Specifically, we encode the sampled coordinate-wise point features into a continuous space, and then align position and features with respect to the continuous coordinate. 

We further hinge on the idea of mixture-of-experts (MoE) to improve segmentation quality. 
Considering our goal is to rectify uncertain coarse predictions, we regard multi-scale representations from the decoder as experts. 
During training, experts are randomly activated for features from multiple blocks of the decoder, and correspondingly the INRs of multi-scale representations are separately parameterized by a group of MLPs that compose a spanning set of the target function class. 
In this way, the INRs are acquired across the multi-block structure while the stochastic experts are specified by the anatomical features at each block.

In summary, our main contributions are as follows: (1) We propose a new implicit neural rendering framework that has fine-grained control of segmentation quality by adaptively composing INRs (\ie, coordinate-wise point features) and rectifying uncertain anatomical regions; (2) We illustrate the advantage of adopting mixture-of-experts that endows the model with better specialization of features maps for improving the performance; (3) Extensive experiments show that our method consistently improves performance compared to 2D and 3D SOTA CNN- and Transformer-based approaches; and (4) Theoretical analysis verifies the expressiveness of our INR-based model. Code is released at \href{https://github.com/charlesyou999648/MORSE}{here}.

\section{Method}
\label{section:method}

\begin{figure}[t]
\centering
\includegraphics[width=0.8\linewidth]{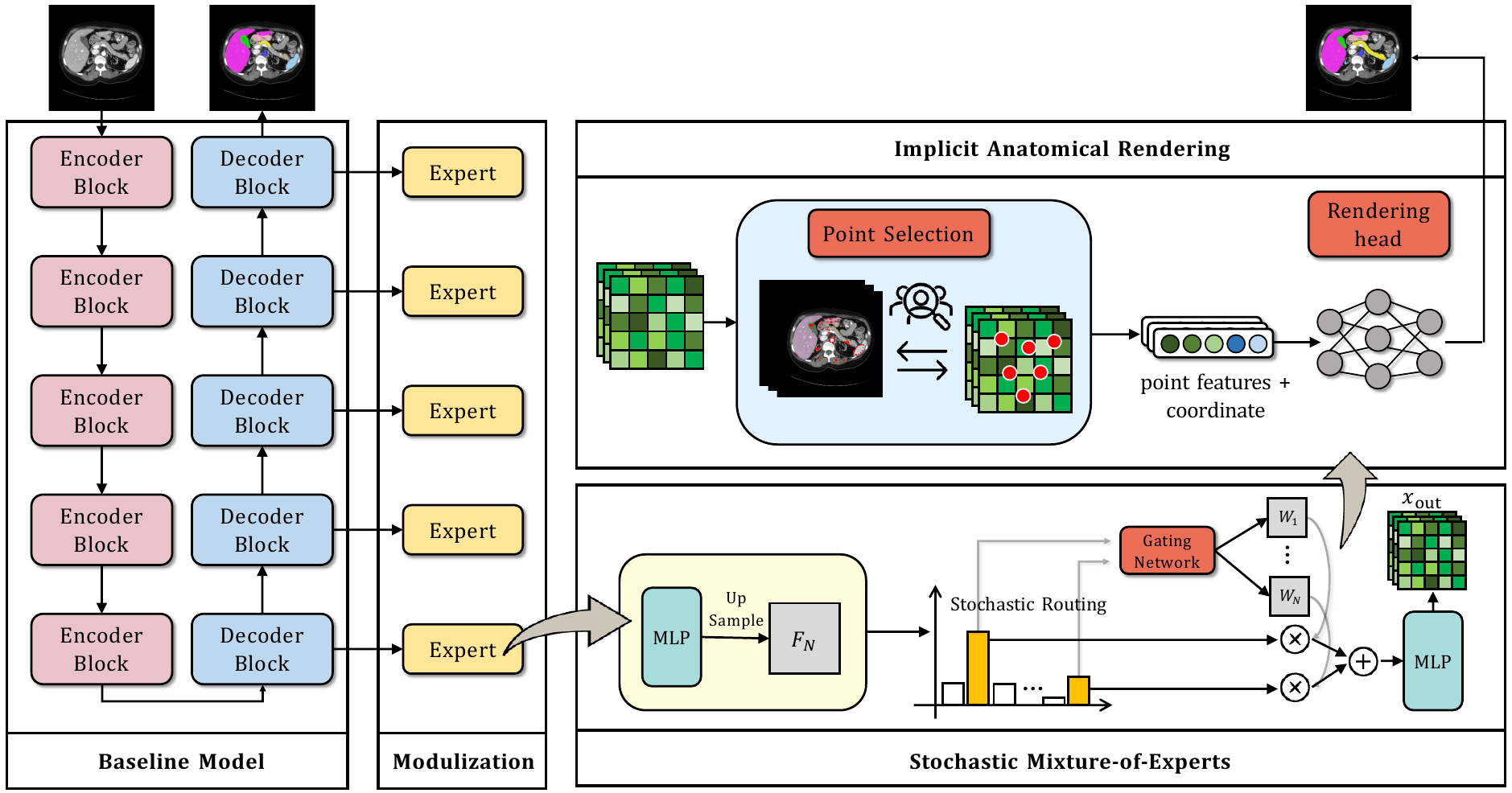}
\vspace{-10pt}
\caption{Illustration of the MORSE pipeline.} 
\label{fig:model}
\vspace{-10pt}
\end{figure}

Let us assume a supervised medical segmentation dataset $\mathcal{D} = \{(x, y)\}$, where each input $x = x_1, x_2, ..., x_T$ is a collection of $T$ 2D/3D scans, and $y$ refers to the ground-truth labels.
Given an input scan ${x}\in\mathbb{R}^{H\times W\times d}$, the goal of medical segmentation is to predict a segmentation map ${\hat{y}}$. Fig.~\ref{fig:model} illustrates the overview of our MORSE. In the following, we first describe our baseline model  $f$ for standard supervised learning, and subsequently present our MORSE.
A baseline segmentation model consists of two main components: (1) encoder module, which generates the multi-scale feature maps such that the model is capable of modeling multi-scale local contexts, and (2) decoder module that makes a prediction $\hat{y}$ using the generated multi-block features of different resolution.
The entire model $M$ is trained end-to-end using the supervised segmentation loss $\mathcal{L}_\text{sup}$ \cite{you2022class} (\ie, equal combination of cross-entropy loss and dice loss).

\vspace{-0.2cm}
\subsection{Stochastic Mixture-of-Experts (SMoE) Module}
\label{sec:stochastic_moe}
\myparagraph{Motivation}
We want a module that encourages inter- and intra-associations across multi-block features. Intuitively, multi-block features should be specified by anatomical features across each block. We posit that due to the specialization-favored nature of MoE, the model will benefit from explicit use of its own anatomical features at each block by learning multi-scale anatomical contexts with adaptively selected experts. In implementation, our SMoE module follows an MoE design  \cite{ou2022patcher}, where it treats features from multiple blocks of the decoder as experts. To mitigate potential overfitting and enable parameter-efficient property, we further randomly activate experts for each input during training. Our approach makes three major departures compared to \cite{ou2022patcher} (\ie, SOTA segmention model): (1) implicitly optimized during training since it greatly trims down the training cost and the model scale; (2) using features from the \textit{decoder} instead of the \textit{encoder} tailored for our refinement goal; and (3) empirically showing that ``self-slimmable'' attribute delivers sufficiently exploited expressiveness of the model.

\myparagraph{Modulization} 
We first use multiple small MLPs with the same size to process different block features and then up-sample the features to the size of the input scans, \ie, $H\times W\times d$. 
With $N$ as the total number of layers (experts) in the decoder, we treat these upsampled features $[\bs{F}_1, \bs{F}_2, ..., \bs{F}_N]$ as expert features. We then train a gating network $\mathcal{G}$ to re-weight the features from activated experts with the trainable weight matrices $[\bs{W}_1, \bs{W}_2, ..., \bs{W}_N]$, where $\bs{W}\in\mathbb{R}^{H\times W\times d}$. 
Specifically, the gating network or router $\mathcal{G}$ outputs these weight matrices satisfying $\sum_i \bs{W}_i = \mathbf{1}^{H\times W\times d}$ using a structure depicted as follows:
\begin{equation}
    \bs{W}_{i} = [\texttt{Softmax}(\texttt{Conv}([\bs{F}_1, \bs{F}_2, ..., \bs{F}_N]))]_i, \quad \textnormal{for} \ \ i \in [N] . 
\end{equation}
The gating network first concatenates all the expert features along channels and uses several convolutional layers to get $\texttt{Conv}([\bs{F}_1, \bs{F}_2, ..., \bs{F}_N]) \in \mathbb{R}^{C \times H\times W\times d\times N}$, where $C$ is the channel dimension.  
A softmax layer is applied over the last dimension (\ie, $N$-expert) to output the final weight maps. 
After that, we feed the resultant output $x_\text{out}$  to another MLP to fuse multi-block expert features. 
Finally, the resultant output $x_\text{out}$ (\ie ~the coarse feature) is given as follows:
\begin{equation}
    x_\text{out} = \texttt{MLP}(\,\sum_{i=1}^N \bs{W}_i\cdot \bs{F}_i\,),
\end{equation}
where $\cdot$ denotes the pixel-wise multiplication, and $x_\text{out}\in \mathbb{R}^{C \times H\times W\times d}$.

\myparagraph{Stochastic Routing} The prior MoE-based model \cite{ou2022patcher} are densely activated. That is, a model needs to access all its parameters to process all inputs. One drawback of such design often comes at the prohibitive training cost. Moreover, the large model size suffers from the representation collapse issue \cite{shazeer2017outrageously}, further limiting the model's performance. 
Our proposed SMoE considers \textit{randomly activated} expert sub-networks to address the issues. 
In implementation, we simply apply standard dropout to multiple experts with a dropping probability $\alpha$. 
For each training iteration, there are dropout masks placed on experts with the probability $\alpha$. 
That is, the omission of experts follows a $\texttt{Bernoulli}(\alpha)$ distribution. 
As for inference, there is no dropout mask and all experts are activated.

\vspace{-0.15cm}
\subsection{Implicit Anatomical Rendering (IAR)}
\label{sec:implicit_render}
The existing methods generally assume that the semantically correlated information and fine anatomical details have been captured and can be used to obtain high-quality segmentation quality. 
However, CNNs inherently operate the discrete signals in a grid of pixels or voxels, which naturally blur the high-frequency anatomical regions, \ie, boundary regions. 
To address such issues, INRs in computer graphics are often used to replace standard discrete representations with continuous functions parameterized by MLPs \cite{tancik2020fourier,sitzmann2020implicit}. 
Our key motivation is that the task of medical segmentation is often framed as a rendering problem that applies implicit neural functions to continuous shape/object/scene representations \cite{park2019deepsdf,sitzmann2020implicit}. 
Inspired by this, we propose an implicit neural rendering framework to further improve segmentation quality, \ie, to adaptively compose coordinate-wise point features and rectify uncertain anatomical regions. 

\myparagraph{Point Selection} 
Given a coarse segmentation map, the rendering head aims at rectifying the uncertain boundary regions. 
A point selection mechanism is thus required to filter out those pixels where the rendering can achieve maximum segmentation quality improvement.
Besides, point selection can significantly reduce computational cost compared to blindly rendering all boundary pixels. 
Therefore, our MORSE selects $N_p$ points for refinement given the coarse segmentation map using an uncertainty-based criterion. 
Specifically, MORSE first uniformly randomly samples $k_{p} N_p$ candidates from all pixels where the hyper-parameter $k_{p} \geq 1$, following \cite{kirillov2020pointrend}. 
Then, based on the coarse segmentation map, MORSE  chooses $\rho N_p$ pixels with the highest uncertainty from these candidates, where $0.5<\rho<1$. 
The uncertainty for a pixel is defined as $\texttt{SecondLargest}(\mathbf{v}) - \texttt{max}(\mathbf{v})$, where $\mathbf{v}$ is the logit vector of that pixel such that the coarse segmentation is given as \texttt{Softmax}$(\mathbf{v})$.
The rest $(1-\rho) N_p$ pixels are sampled uniformly from all the remaining pixels. 
This mechanism ensures the selected points contain a large portion of points with uncertain segmentation which require refinement. 

\myparagraph{Positional Encoding}
It is well-known that neural networks can be cast as universal function approximators, but they are inferior to high-frequency signals due to their limited learning power \cite{rahaman2019spectral,mildenhall2021nerf}. Unlike \cite{kirillov2020pointrend}, we explore using the encoded positional information to capture high-frequency signals, which echoes our theoretical findings in Appendix~\ref{sec:analysis}.
Specifically, for a coordinate-based point $(x, y) \in [H]\times [W]$, the positional encoding function is given as:
\begin{align}\label{eq:positional encoding}
    \psi(x, y) & = [\sin(2\pi(w_1 \tilde{x}+v_1 \tilde{y})), \cdots , \sin(2\pi(w_{L} \tilde{x}+v_{L} \tilde{y})), \notag
    \\ & \qquad \cos(2\pi(w_1 \tilde{x}+v_1 \tilde{y})), \cdots , \cos(2\pi(w_{L} \tilde+v_{L} \tilde{y}))], 
\end{align}where $\tilde{x} = 2x/H-1$ and $\tilde{y} = 2y/W-1$ are the standardized coordinates with values in between $[-1,1]$.
The frequency $\{w_i, v_i\}_{i=1}^L$ are trainable parameters with Gaussian random initialization, where we set $L = 128$~\cite{cheng2022pointly}. 
For each selected point, its position encoding will then be concatenated with the coarse features of that point (\ie, $x_{\textnormal{out}}$ defined in Sec.~\ref{sec:stochastic_moe}), to output the fine-grained features. 

\myparagraph{Rendering Head} 
The fine-grained features are then fed to the rendering head whose goal is to rectify the uncertain predictions with respect to these selected points. Inspired by \cite{kirillov2020pointrend}, the rendering head adopts 3-layer MLPs design.
Since the rendering head is designed to rectify the class label of the selected points, it is trained using the standard cross-entropy loss $\mathcal{L}_\text{rend}$. 

\myparagraph{Adaptive Weight Adjustment} 
Instead of directly leveraging pre-trained weights, it is more desirable to train the model \textit{from scratch} in an  \textit{end-to-end} way. For instance, we empirically observe that directly using coarse masks by pre-trained weights to modify unclear anatomical regions might lead to suboptimal results (See Sec. \ref{subsection:comparision}). Thus, we propose to modify the importance of $\mathcal{L}_\text{rend}$ as:
\begin{equation}
    \lambda_t = \lambda_{\textnormal{rend}} \cdot \left[\mathbbm{1}\{t >T/2\} \cdot\left(\frac{t-T/2}{T} \right) \right],
\end{equation}
where $t$ is the index of the iteration, $T$ denotes the total number of iterations, and $\mathbbm{1}\{\cdot\}$ denotes the indicator function. 

\myparagraph{Training Objective} As such, the model is trained in an \textit{end-to-end} manner using total loss $\mathcal{L}_\text{total}=\mathcal{L}_\text{sup}+\lambda_t\times\mathcal{L}_\text{rend}$.

\section{Experiments}
\label{section:exp}
\begin{table}[t]
\centering
\caption{Quantitative comparisons for multi-organ segmentation on the Synapse multi-organ CT dataset. The best results are indicated in \tf{bold}.}
\label{table:synapse_main}
\vspace{-8pt}
\begin{adjustbox}{width=\textwidth}
\begin{tabular}{@{}l|cccc|cccccccc@{}}
\toprule
\multirow{2}{*}{Method}          
& \multicolumn{4}{c|}{Average}      
& \multirow{2}{*}{Aorta} 
& \multirow{2}{*}{Gallbladder} 
& \multirow{2}{*}{Kidney (L)}
& \multirow{2}{*}{Kidney (R)} 
& \multirow{2}{*}{Liver} 
& \multirow{2}{*}{Pancreas} 
& \multirow{2}{*}{Spleen} 
& \multirow{2}{*}{Stomach}       \\
& DSC\,$\uparrow$ 
& Jaccard\,$\uparrow$ 
& 95HD\,$\downarrow$ 
& ASD\,$\downarrow$             
&          
&         
&             
&            
&             
&           
&             
& 
\\ \midrule
UNet (Baseline) \cite{ronneberger2015u}         
& 70.11        
& 59.39            
& 44.69         
& 14.41              
& 84.00             
& 56.70           
& 72.41            
& 62.64          
& 86.98 
& 48.73   
& 81.48       
& 67.96   
\\
\ + PointRend \cite{kirillov2020pointrend}                
& 71.52        
& 61.34            
& 43.19         
& 13.70              
& 85.74             
& 57.14           
& 75.42            
& 63.27          
& 87.32
& 50.16  
& 81.82       
& 71.29           
\\
\ + Implicit PointRend \cite{cheng2022pointly}                
& 67.33        
& 59.73            
& 52.44         
& 22.15            
& 76.32             
& 51.99           
& 70.28            
& 70.36          
& 81.69 
& 43.77   
& 77.18       
& 67.05
\\
\ + Ours (MoE)                
& 72.83        
& 62.64            
& 40.44         
& 13.15              
& 86.11             
& 59.51           
& 75.81            
& 67.10          
& 87.82 
& 52.11   
& 83.48       
& 70.86
\\
\ + Ours (SMoE)                
& 74.86       
& 64.94            
& 37.69         
& 12.66              
& 86.39             
& 63.99           
& 77.96            
& 68.93          
& 88.88 
& 53.62   
& 86.12       
& 72.98
\\
\ + Ours (IAR)                
& 73.11        
& 62.98            
& 34.01         
& 12.67              
& 86.28             
& 60.25           
& 76.58            
& 65.34          
& 88.32 
& 52.12   
& 83.47       
& 72.51
\\
\ + Ours (IAR+MoE)                
& 75.37        
& 65.65            
& 33.34         
& 11.43              
& 87.00             
& 64.45           
& 78.14            
& 70.13          
& 89.32 
& 52.33   
& 85.20       
& 76.40
\\
\ + Ours (MORSE)                
& \textbf{76.59}        
& \textbf{66.97}           
& \textbf{32.00}        
& \textbf{10.67}             
& \textbf{87.28}            
& \textbf{64.73}          
& \textbf{80.58}           
& \textbf{71.87}         
& \textbf{90.04}
& \textbf{54.60}  
& \textbf{86.67}      
& \textbf{76.93}
\\
\midrule
TransUnet (Baseline) \cite{chen2021transunet}         
& {77.49}        
& {64.78} 
& {31.69}  
& {8.46}       
& 87.23        
& 63.13          
& 81.87          
& 77.02          
& 94.08      
& 55.86         
& 85.08      
& 75.62  
\\
\ + PointRend \cite{kirillov2020pointrend}                
& 78.30        
& 65.88            
& 34.17         
& 8.62         
& 87.93             
& 63.96           
& 83.47            
& 77.23          
& 94.86 
& 56.45   
& 85.76       
& 76.75           
\\
\ + Implicit PointRend \cite{cheng2022pointly}                
& 71.92       
& 60.62        
& 41.42     
& 18.55          
& 78.39         
& 61.64          
& 79.59        
& 73.20 
& 89.61
& 50.01
& 80.17   
& 62.75
\\
\ + Ours (MoE)                
& 77.85        
& 65.30            
& 32.75         
& 7.90              
& 87.40             
& 63.46           
& 82.34            
& 77.88          
& 94.14 
& 56.12   
& 85.24       
& 76.25
\\
\ + Ours (SMoE)                
& 78.68        
& 65.98            
& 31.86         
& 7.00              
& 87.60             
& 66.21           
& 82.62            
& 78.12          
& 94.88 
& 57.59   
& 85.97       
& 76.48
\\
\ + Ours (IAR)                
& 79.37        
& 66.50            
& 30.13         
& 7.25              
& 88.63             
& 66.76           
& 83.70            
& 79.50          
& 95.26 
& 57.10   
& 86.90       
& 77.10
\\
\ + Ours (IAR+MoE)                
& 79.60       
& 66.99            
& 27.59         
& 6.54              
& 88.73             
& 66.83           
& 83.85            
& 80.19          
& 95.98 
& 57.12   
& 86.92       
& 77.21
\\
\ + Ours (MORSE)                
& \textbf{80.85}       
& \textbf{68.53}           
& \textbf{26.61}         
& \textbf{6.46}            
& \textbf{88.92}            
& \textbf{67.53}          
& \textbf{84.83}           
& \textbf{81.68}         
& \textbf{96.83}
& \textbf{59.70}  
& \textbf{87.73}      
& \textbf{79.58}
\\
\bottomrule
\end{tabular}
\end{adjustbox}
\vspace{-10pt}
\end{table}

\myparagraph{Dataset}
We evaluate the models on two important medical segmentation tasks. (1) \textbf{Synapse multi-organ segmentation}\footnote{\url{https://www.synapse.org/\#!Synapse:syn3193805/wiki/217789}}:
Synapse multi-organ segmentation dataset contains 30 abdominal CT scans with 3779 axial contrast-enhanced abdominal clinical CT images in total. Each volume scan has variable volume sizes $512\times 512 \times 85$\,$\sim$\,$512\times 512 \times 198$ with a voxel spatial resolution of $([0.54\!\sim\!0.54]\times[0.98\!\sim\!0.98]\times[2.5\!\sim\!5.0])$mm$^3$. For a fair comparison, the data split\footnote{\url{https://github.com/Beckschen/TransUNet/tree/main/lists/lists_Synapse}} is fixed with 18 (2211 axial slices) and 12 patients' scans for training and testing, respectively. The entire dataset has a high diversity of aorta, gallbladder, spleen, left kidney, right kidney, liver, pancreas, spleen, and stomach. \\
(2) \textbf{Liver segmentation}: Multi-phasic MRI (MP-MRI) dataset is an in-house dataset including 20 patients, each including T1 weighted DCE-MRI images at three-time phases (\ie, pre-contrast, arterial, and venous). Here, our evaluation is conducted via 5-fold cross-validation on the 60 scans. For each fold, the training and testing data includes 48 and 12 cases, respectively.

\myparagraph{Implementation Details}
We use AdamW optimizer \cite{loshchilov2018decoupled} with an initial learning rate $5e^{-4}$, and adopt a polynomial-decay learning rate schedule for both datasets. We train each model for 30K iterations. For Synapse, we adopt the input resolution as 256$\times$256 and the batch size is 4. For MP-MRI, we randomly crop 96$\times$96$\times$96 patches and the batch size is 2. For SMoE, following \cite{ou2022patcher}, all the MLPs have hidden dimensions $[256,256]$ with ReLU activations, the dimension of expert features $[\bs{F}_1, \bs{F}_2, ..., \bs{F}_N]$ are 256. We empirically set $\alpha$ as 0.7. Following \cite{kirillov2020pointrend}, $N_p$ is set as 2048, and 8192 for training and testing, respectively, and $k_{p}$, $\rho$ are 3, 0.75.
We follow the same gating network design \cite{ou2022patcher}, which includes four $3\times 3$ convolutional layers with channels $[256,256,256,N]$ and ReLU activations. $\lambda_{\text{rend}}$ are set to 0.1. 
We adopt four representative models, including UNet \cite{ronneberger2015u}, TransUnet \cite{chen2021transunet}, 3D-UNet \cite{cciccek20163d}, UNETR \cite{hatamizadeh2021unetr}. Specifically, we set $N$ for UNet \cite{ronneberger2015u}, TransUnet \cite{chen2021transunet}, 3D-UNet \cite{cciccek20163d}, UNETR \cite{hatamizadeh2021unetr} with 5, 3, 3, 3, respectively.
We also use Dice coefficient (DSC), Jaccard, 95\% Hausdorff Distance (95HD), and Average Surface Distance (ASD) to evaluate 3D results. We conduct all experiments in the same environments with fixed random seeds (Hardware: Single NVIDIA RTX A6000 GPU; Software: PyTorch 1.12.1+cu116, and Python 3.9.7).

\begin{table}[t]
\centering
\caption{Quantitative comparisons for liver segmentation on the Multi-phasic MRI dataset. The best results are indicated in \tf{bold}.}
\label{table:liver_main}
\vspace{-8pt}
\begin{adjustbox}{width=0.97\textwidth}
\begin{tabular}{@{}l|cccc|l|cccc@{}}
\toprule
\multirow{2}{*}{Method}          
& 
\multicolumn{4}{c|}{Average}         
& \multirow{2}{*}{Method}          
& \multicolumn{4}{c}{Average}    
\\ 
& DSC\,$\uparrow$ 
& Jaccard\,$\uparrow$ 
& 95HD\,$\downarrow$ 
& ASD\,$\downarrow$    
&
& DSC\,$\uparrow$ 
& Jaccard\,$\uparrow$ 
& 95HD\,$\downarrow$ 
& ASD\,$\downarrow$    
\\ \midrule
3D-UNet (Baseline) \cite{cciccek20163d}         
& 89.19       
& 81.21
& 34.97         
& 10.63       
& UNETR (Baseline) \cite{hatamizadeh2021unetr}         
& {89.95}        
& {82.17} 
& {24.64}  
& {6.04}       
\\
\ + PointRend \cite{kirillov2020pointrend}                
& 89.55   
& 81.80            
& 30.88         
& 10.12         
& + PointRend \cite{kirillov2020pointrend}                
& 90.49       
& 82.36            
& 21.06         
& 5.59   
\\
\ + Implicit PointRend \cite{cheng2022pointly}                
& 88.01 
& 79.83        
& 37.55     
& 12.86        
& + Implicit PointRend \cite{cheng2022pointly}                
& 88.72      
& 80.18        
& 26.63     
& 10.58    
\\
\ + Ours (MoE)                
& 89.81        
& 82.06            
& 29.96        
& 10.15     
& + Ours (MoE)                
& 90.70        
& 82.80            
& 15.31         
& 5.93   
\\
\ + Ours (SMoE)                
& 90.16       
& 82.28            
& 28.36         
& 9.79        
& + Ours (SMoE)                
& 91.02        
& 83.29            
& 15.12         
& 5.64   
\\
\ + Ours (IAR)                
& 91.22        
& 83.30       
& 27.84         
& 8.89       
& + Ours (IAR)                
& 91.63     
& 83.83            
& 14.25         
& 4.99   
\\
\ + Ours (IAR+MoE)                
& 92.77        
& 83.94            
& 26.57         
& 7.51       
& + Ours (IAR+MoE)                
& 93.01      
& 84.70            
& 13.29         
& 4.84  
\\
\ + Ours (MORSE)                
& \textbf{93.59}        
& \textbf{84.62}           
& \textbf{19.61}        
& \textbf{6.57}   
& + Ours (MORSE)                
& \textbf{93.85}       
& \textbf{85.53}           
& \textbf{12.33}         
& \textbf{4.38} 
\\
\bottomrule
\end{tabular}
\end{adjustbox}
\vspace{-10pt}
\end{table}

\vspace{-0.1cm}
\subsection{Comparison with State-of-the-Art Methods}
\label{subsection:comparision}
We adopt classical CNN- and transformer-based models, \ie, 2D-based \{UNet \cite{ronneberger2015u}, TransUnet \cite{chen2021transunet}\} and 3D-based \{3D-UNet \cite{cciccek20163d}, UNETR \cite{hatamizadeh2021unetr}\}, and train them on \{2D Synapse, 3D MP-MRI\} in an end-to-end manner\,\footnote{All comparison experiments are using their released code.}.

\myparagraph{Main Results}
The results for 2D synapse multi-organ segmentation and 3D liver segmentation are shown in Tables \ref{table:synapse_main} and \ref{table:liver_main}, respectively. The following observations can be drawn: (1) Our MORSE demonstrates superior performance compared to all other training algorithms. Specifically, Compared to UNet, TransUnet, 3D-UNet, and UNETR baselines, our MORSE with all experts selected obtains 3.36\%$\sim$6.48\% improvements in Dice across two segmentation tasks. It validates the superiority of our proposed MORSE. (2) The stochastic routing policy shows consistent performance benefits across all four network backbones on 2D and 3D settings. Specifically, we can observe that our SMoE framework improves all the baselines, which is within expectation since our model is implicitly ``optimized'' given evolved features. 
(3) As is shown, we can observe that IAR consistently outperforms PointRend across all the baselines (\ie, UNet, TransUnet, 3D-UNet, and UNETR) and obtain \{1.59\%, 1.07\%, 2.03\%, 1.14\%\} performance boosts on two segmentation tasks, highlighting the effectiveness of our proposal in INRs.
(4) With Implicit PointRend \cite{cheng2022pointly} equipped, all the models' performances drop. We find: adding Implicit PointRend leads to significant performance drops of -2.78\%, -5.57\%, -1.18\%, and -1.23\% improvements, compared with the SOTA baselines (\ie, UNet, TransUnet, 3D-UNet, and UNETR) on two segmentation tasks, respectively. Importantly, we find that: \cite{cheng2022pointly} utilizes INRs for producing different parameters of the point head for each object with point-level supervision. As this implicit function does not directly optimize the anatomical regions, we attribute this drop to the introduction of additional noise during training, which leads to the representation collapse. This further verifies the effectiveness of our proposed IAR. In Appendix Figs. \ref{fig:vis_synpase} and \ref{fig:vis_jhu}, we provide visual comparisons from various models. We can observe that MORSE yields sharper and more accurate boundary predictions compared to all the other training algorithms.

\myparagraph{Visualization of IAR Modules}
To better understand the IAR module, we visualize the point features on the coarse prediction and refined prediction after the IAR module in Appendix Fig. \ref{fig:vis_bound}. As is shown, we can see that IAR help rectify the uncertain anatomical regions for improving segmentation quality. 

\begin{table}
\parbox{.435\linewidth}{
\vspace{-0.2mm}
\caption{Effect of stochastic rate $\alpha$ and expert number $N$.}\label{tab:ablation_size}
\vspace{-3.5mm}
\resizebox{\linewidth}{!}
{
\begin{tabular}{@{\hskip 1mm}cccccc@{\hskip 3mm}ccccccc@{\hskip 1mm}}
\toprule
$\alpha$ && {DSC{[}\%{]}$\uparrow$} && ASD{[}voxel{]$\downarrow$} && $N$ && {DSC{[}\%{]}$\uparrow$} && ASD{[}voxel{]$\downarrow$}\\ 
\cmidrule(l{0mm}r{0.5mm}){1-5} \cmidrule(l{0.3mm}r{0mm}){7-11}
0.1  && 75.41 && 11.96 && 1 (No MoE)  &&75.11 && 11.67 \\
0.2 && 75.68 && 11.99 && 2  && 75.63 && 11.49\\
0.5 && 76.06 && \textbf{10.43} && 3  && 75.82 && 11.34 \\
0.7  && \textbf{76.59} && 10.67 && 4  && 76.16 && 11.06 \\
0.9  && 74.16 && 11.32 && 5  && \textbf{76.59} && \textbf{10.67} \\
\bottomrule
\end{tabular}
}
}
\hfill
\parbox{.53\linewidth}{
\caption{Ablation studies of the Adaptive Weight Adjustment (AWA).}
\vspace{-3.5mm}
\label{table:ablation_ats}
\resizebox{\linewidth}{!}
{
\begin{tabular}{@{\hskip 1mm}lcccccc@{\hskip 1mm}}
\toprule
Method  && {DSC{[}\%{]}$\uparrow$} && ASD{[}voxel{]$\downarrow$}\\ 
\midrule
w/o AWA \& train w/ $\mathcal{L}_\text{rend}$ from scratch && 70.56 && 14.89 \\
w/o AWA \& train w/ $\mathcal{L}_\text{rend}$ in $\frac{T}{2}$ && 75.42 && 12.00 \\
w/ AWA && \textbf{76.59} && \textbf{10.67} \\
\bottomrule
\end{tabular}
}
}
\vspace{-10mm}
\end{table}

\subsection{Ablation Study}
We first investigate our MORSE equipped with UNet by varying $\alpha$ (\ie, stochastic rate) and $N$ (\ie, experts) on Synapse. The comparison results of $\alpha$ and $N$ are reported in Table \ref{tab:ablation_size}. We find that using $\alpha=0.7$ performs the best when the expert capacity is $N=5$. Similarly, when reducing the expert number, the performance also drops considerably. This shows our hyperparameter settings are optimal.

Moreover, we conduct experiments to study the importance of Adaptive Weight Adjustment (AWA). We see that: (1) Disabling AWA and training $\mathcal{L}_\text{rend}$ from scratch causes unsatisfied performance, as echoed in \cite{kirillov2020pointrend}. (2) Introducing AWA shows a consistent advantage compared to the other. This demonstrates the importance of the Adaptive Weight Adjustment.
\section{Conclusion}
In this paper, we proposed MORSE, a new implicit neural rendering framework that has fine-grained control of segmentation quality by adaptively composing coordinate-wise point features and rectifying uncertain anatomical regions. We also demonstrate the advantage of leveraging mixture-of-experts that enables the model with better specialization of features maps for improving the performance.
Extensive empirical studies across various network backbones and datasets, consistently show the effectiveness of the proposed MORSE. Theoretical analysis further uncovers the expressiveness of our INR-based model.

%
%
%
\bibliographystyle{splncs04}
\bibliography{mybibliography}

\clearpage
\appendix 
\section*{Appendix}

\begin{figure}[ht]
\vspace{-10mm}
\centering
\includegraphics[width=0.65\linewidth]{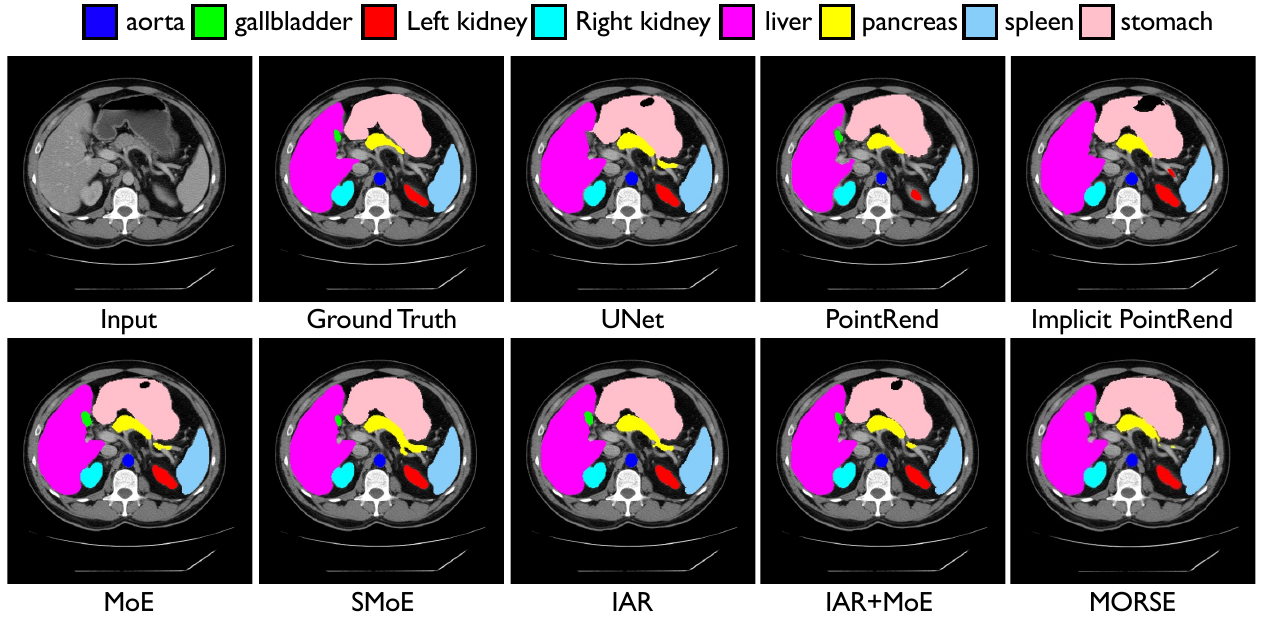}
\vspace{-12pt}
\caption{Visualization results on Synpase. MORSE yields more accurate predictions, especially for small regions.} 
\label{fig:vis_synpase}
\vspace{-15pt}
\end{figure}

\begin{figure}[h]
\vspace{-10mm}
\centering
\includegraphics[width=0.65\linewidth]{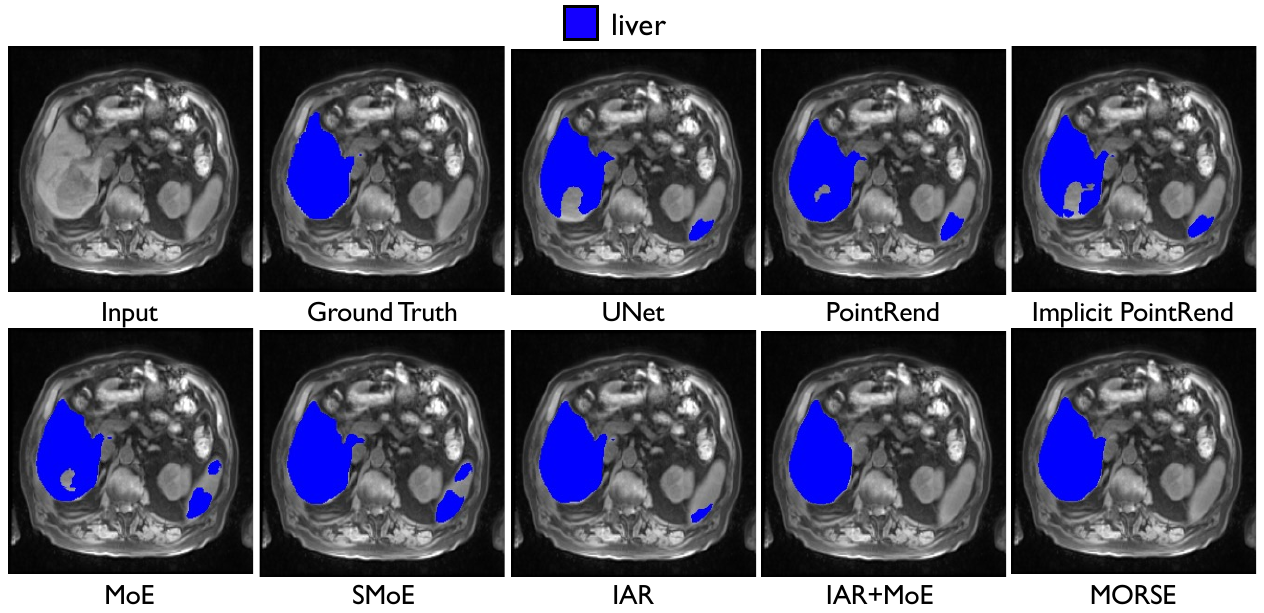}
\vspace{-12pt}
\caption{Visualization results on MP-MRI. MORSE outputs more accurate segmentation results, especially for boundaries.} 
\label{fig:vis_jhu}
\vspace{-15pt}
\end{figure}

\begin{figure}[h]
\vspace{-10mm}
\centering
\includegraphics[width=0.6\linewidth]{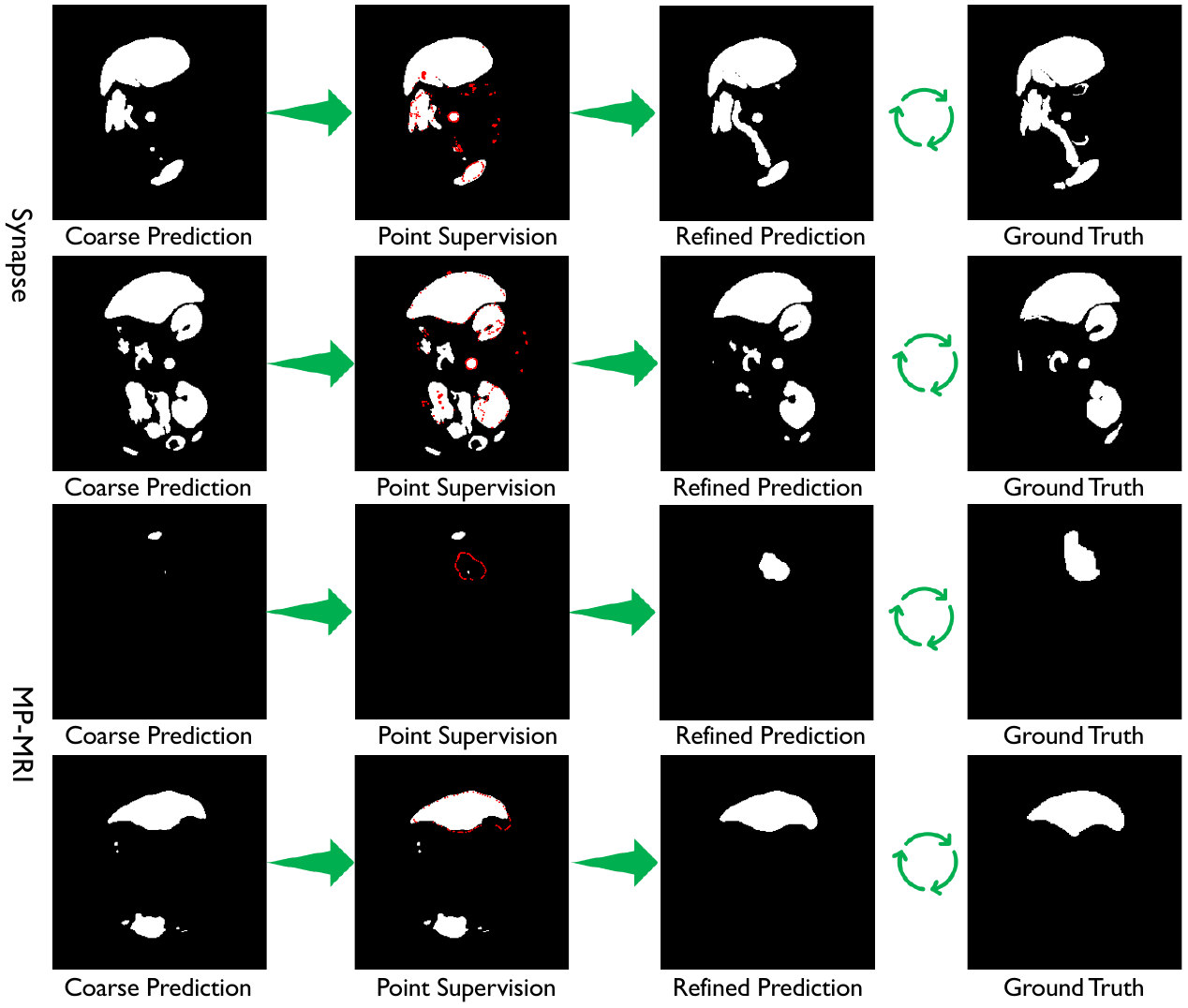}
\vspace{-12pt}
\caption{Visualization of sampled point location. As is shown, IAR significantly improves the segmentation quality.} 
\label{fig:vis_bound}
\vspace{-25pt}
\end{figure}

\section{Theoretical Analysis}\label{sec:analysis}

In this section, we theoretically analyze the expressiveness of MORSE, and demonstrate the approximation power of INR features combined with MLPs. 

We first introduce the kernel method which is  commonly used in analyzing neural networks. 
We define the kernel function $\kappa$: $\mathbb{R}^{d} \times \mathbb{R}^{d} \to \mathbb{R}$, which is a symmetric function measuring the similarity between two vectors in $\mathbb{R}^d$. 
Kernel functions are used to approximate unknown functions from data.
Specifically, given a training dataset $\{\mathbf{x}_i, \mathbf{y}_i\}_{i=1}^n$ where $\mathbf{y}_i = \mu(\mathbf{x}_i)$ for some unknown function $\mu(\cdot)$, an estimate $\hat{\mu}$ can be constructed using kernel function as $\hat{\mu}(\mathbf{x}) = \sum_{i=1}^n (\mathbf{K}^{-1} \mathbf{Y})_i \kappa(\mathbf{x}_i, \mathbf{x})$ , where $\mathbf{K}$ is the $n\times n$ matrix with $\mathbf{K}_{i,j} = \kappa(\mathbf{x}_i, \mathbf{x}_j)$, and $\mathbf{Y} = [\mathbf{y}_1, \cdots, \mathbf{y}_n ]^\top$. 

The kernel method is related to MLPs through a kernel function called Neural Tangent Kernel \cite{jacot2018neural}. 
For an MLP $\Pi_\theta$ with trainable parameters $\theta$, it has been shown that, under certain conditions, the model $\Pi_\theta$ trained with stochastic gradient descent will converge to the estimate generated by the kernel method through Neural Tangent Kernel, which is defined as $\kappa_{\text{NTK}}(\mathbf{x}_i, \mathbf{x}_j) = \mathbb{E}_{\theta \sim \mathcal{N}} \left\langle \frac{\partial \Pi_{\theta}(\mathbf{x}_i) }{\partial \theta} , \frac{\partial \Pi_{\theta}(\mathbf{x}_j) }{\partial \theta} \right\rangle$.
Importantly, for $\mathbf{x}_i, \ \mathbf{x}_j$ on the unit sphere, NTK is an inner product kernel, \ie, 
\begin{align}\label{eq:ntk}
    \kappa_{\text{NTK}}(\mathbf{x}_i, \mathbf{x}_j) = k_{\text{NTK}}(\mathbf{x}_i \cdot \mathbf{x}_j),
\end{align} for some function $k_{\text{NTK}}$.

With the kernel function explained, we now show how positional encoding helps with expressiveness.  
By construction, the positional encoding maps pixel coordinates into sinusoidal vectors. This is closely related to random Fourier features which are provably able to approximate the family of shift-invariant kernel functions \cite{rahimi2007random}.  
Specifically, a kernel function $\kappa$ is shift-invariant if  for any $\mathbf{x}_1, \ \mathbf{x}_2 \in \mathbb{R}^d$, it holds that $\kappa(\mathbf{x}_1, \mathbf{x}_2) = \kappa(\mathbf{x}_1- \mathbf{x}_2)$ with a slight abuse of notation. 
In other words, the value of $\kappa(\mathbf{x}_1, \mathbf{x}_2)$ only depends on the difference $\mathbf{x}_1- \mathbf{x}_2$. 
It is clear that shift-invariance is an ideal property in imaging tasks. 

To understand how the Fourier features generated by positional encoding can approximate shift-invariant kernel, we rewrite positional encoding (Eqn. \ref{eq:positional encoding}) as:
\begin{align*}
    \psi(\mathbf{x}) = [\sin(2\pi \mathbf{w}_1\cdot\mathbf{x}), \cdots, \ \sin(2\pi \mathbf{w}_L\cdot\mathbf{x}), \cos(2\pi \mathbf{w}_1\cdot\mathbf{x}), \cdots, \ \cos(2\pi \mathbf{w}_L\cdot\mathbf{x})] , 
\end{align*} where $\mathbf{x} = (\tilde{x}, \tilde{y})$ , and $\mathbf{w}_i = [w_i, v_i]$ for $i=1, \cdots, L$. 
Given any shift-invariant kernel $\kappa(\mathbf{x}_1, \mathbf{x}_2) = \kappa(\mathbf{x}_1 - \mathbf{x}_2)$, we define a distribution  $\mathcal{P}$ over $\mathbf{w}$ as $\mathcal{P}(\mathbf{w}) = \frac{1}{2\pi} \int e^{- 2\pi i \mathbf{w}^\top \mathbf{x} } \kappa(\mathbf{x}) d \mathbf{x}$,
which is the Fourier transform of the kernel $\kappa$. 
Suppose that $\mathbf{w}_1, \cdots, \mathbf{w}_L$ are $i.i.d.$ samples from $\mathcal{P}$. 
Then it holds that \cite{rahimi2007random}:
\begin{align*}
    \textnormal{Pr}\left[ \sup_{\mathbf{x}_1, \mathbf{x}_2}\left| \frac{1}{L}\psi(\mathbf{x}_1)^\top\psi(\mathbf{x}_2) - \kappa(\mathbf{x}_1,\mathbf{x}_2)\right| \geq \epsilon\right] \leq 2^8 \frac{\sigma_{\mathcal{P}}^2}{\epsilon^2} \exp\left( -\frac{L \epsilon^2}{16}\right),
\end{align*} where $ \sigma_{\mathcal{P}}^2 = \mathbb{E}_{\mathcal{P}} (\mathbf{w}^\top \mathbf{w})$. 
The above result indicates that with high probability, any shift-invariant kernel function can be approximated with Fourier features. 
Therefore, this demonstrates the expressive power of INR features. 

Finally, we show that $\psi(\cdot)$ combined with the MLP $\Pi_\theta$ forms a shift-invariant kernel. 
We define the positional encoding kernel $\kappa_{\text{pe}}$ as $\kappa_{\text{pe}}(\mathbf{x}_1, \mathbf{x}_2) = \psi(\mathbf{x}_1)^\top \psi(\mathbf{x}_2)$. 
It can be shown that $\kappa_{\text{pe}}(\mathbf{x}_1, \mathbf{x}_2) = \sum_{j=1}^L \cos(2\pi \mathbf{w}_j \cdot (\mathbf{x}_1 - \mathbf{x}_2) )$ \cite{tancik2020fourier}, which is shift-invariant.  
Combining with Eqn.~\ref{eq:ntk}, our positional encoding combined followed by MLP approximately equals to $k_{\text{NTK}}(\kappa_{\text{pe}}(\mathbf{x}_1 - \mathbf{x}_2))$, which is a shift-invariant kernel. 
This demonstrates the expressiveness of MORSE.

\end{document}